\documentclass[conference]{IEEEtran}
\IEEEoverridecommandlockouts
\usepackage{multirow}
\usepackage{CJKutf8}
\usepackage{listings}
\usepackage{graphicx}
\usepackage{cite}
\usepackage{amsmath,amssymb,amsfonts}
\usepackage{algorithmic}
\usepackage{graphicx}
\usepackage{textcomp}
\usepackage{xcolor}
\def\BibTeX{{\rm B\kern-.05em{\sc i\kern-.025em b}\kern-.08em
    T\kern-.1667em\lower.7ex\hbox{E}\kern-.125emX}}
\begin{document}

\title{Full-text Error Correction for Chinese Speech Recognition with Large Language Model}

\author{\IEEEauthorblockN{Zhiyuan Tang$^1$, Dong Wang$^2$, Shen Huang$^1$, Shidong Shang$^1$}\\
$^1$Tencent Ethereal Audio Lab, Tencent, China\\
$^2$Center for Speech and Language Technologies, BNRist, Tsinghua University, China\\
atomtang@tencent.com, wangdong99@mails.tsinghua.edu.cn
}

\maketitle

\begin{abstract}
Large Language Models (LLMs) have demonstrated substantial potential for error correction in Automatic Speech Recognition (ASR). However, most research focuses on utterances from short-duration speech recordings, which are the predominant form of speech data for supervised ASR training. This paper investigates the effectiveness of LLMs for error correction in full-text generated by ASR systems from longer speech recordings, such as transcripts from podcasts, news broadcasts, and meetings.
First, we develop a Chinese dataset for full-text error correction, named ChFT, utilizing a pipeline that involves text-to-speech synthesis, ASR, and error-correction pair extractor. This dataset enables us to correct errors across contexts, including both full-text and segment, and to address a broader range of error types, such as punctuation restoration and inverse text normalization, thus making the correction process comprehensive.
Second, we fine-tune a pre-trained LLM on the constructed dataset using a diverse set of prompts and target formats, and evaluate its performance on full-text error correction. Specifically, we design prompts based on full-text and segment, considering various output formats, such as directly corrected text and JSON-based error-correction pairs.
Through various test settings, including homogeneous, up-to-date, and hard test sets, we find that the fine-tuned LLMs perform well in the full-text setting with different prompts, each presenting its own strengths and weaknesses. This establishes a promising baseline for further research.
The dataset is available on the website\footnote{https://huggingface.co/datasets/tzyll/ChFT}.

\end{abstract}


\section{Introduction}
Automatic Speech Recognition (ASR) systems are widely used in a variety of applications, including voice search, voice command, and transcription services. However, the efficacy of ASR can be significantly affected by various factors, such as background noise, speaker accents, and audio signal fidelity. Errors in the ASR output, particularly in challenging environments, can negatively impact the functionality of downstream applications. Therefore, implementing subsequent error correction processes is vital for enhancing the accuracy of ASR outputs.

The use of a language model (LM) to rescore the N-best hypotheses from ASR beam search decoding is a common technique to identify the candidate with the lowest perplexity, as demonstrated in various studies~\cite{mikolov2010recurrent,arisoy2015bidirectional,shin2019effective,yang2021multi,yu2023low}. However, such LM rescoring merely selects the optimal candidate, thereby neglecting the valuable information contained in the remaining hypotheses. A potentially more advantageous strategy involves merging the N-best hypotheses to generate a new prediction, which is expected to be more accurate than the initial candidates~\cite{guo2019spelling,hu2020deliberation,leng2021fastcorrect,hu2022improving,ma2023n,hu2023scaling}.

Recently, large language models (LLMs) have begun leveraging their capacity for understanding language to assist in error correction in a generative style~\cite{chen2023generative,chen2024hyporadise,tang24_interspeech}. Specifically, these methods are designed to generate correct transcription directly from N-best hypotheses produced by ASR systems for a given audio input.

However, due to the nature of the training data for ASR systems, most existing work on generative error correction has focused on single sentences from transcripts, which are limited in their ability to capture the context of the entire conversation or full-text of a document. Moreover, utterance-level error correction is more computationally expensive as it requires generating multiple hypotheses for each utterance, which can be time-consuming and resource-intensive.
In this paper, we propose a novel and comprehensive benchmark specifically designed for generative error correction in full-text documents. This benchmark aims to thoroughly explore and evaluate the potential of LLMs in identifying and correcting a wide range of errors within full-text. 
By full-text, we mean the entire text of a document, such as an article, a news report, or a conversation transcript. This also allows for error types from punctuation restoration and inverse text normalization (ITN), making the task comprehensive.
Specifically, under the relatively long text, we explore the potential of LLMs to correct errors in two dimensions of the text, including full-text and segment composed of multiple sentences. We also design two kinds of output formats, i.e., direct output, which is a direct and obvious correction of the input text, and JSON of error-correction pairs, which is more compact and controllable for subsequent processing such as error detection and then correction.

The remainder of the paper is structured as follows. Section~\ref{sec:dataset} introduces the Chinese full-text error correction dataset, and the pipeline for its construction and fine-tuning dataset preparation.
Section~\ref{sec:prompt} details the prompt design and the output format for different text types.
Section~\ref{sec:experiments} outlines the experimental framework, findings and limitations.
The paper is concluded in Section~\ref{sec:conclusion}.

\begin{figure*}[htbp]
    \centering
    \includegraphics[trim=0 625 0 40,clip,width=\linewidth]{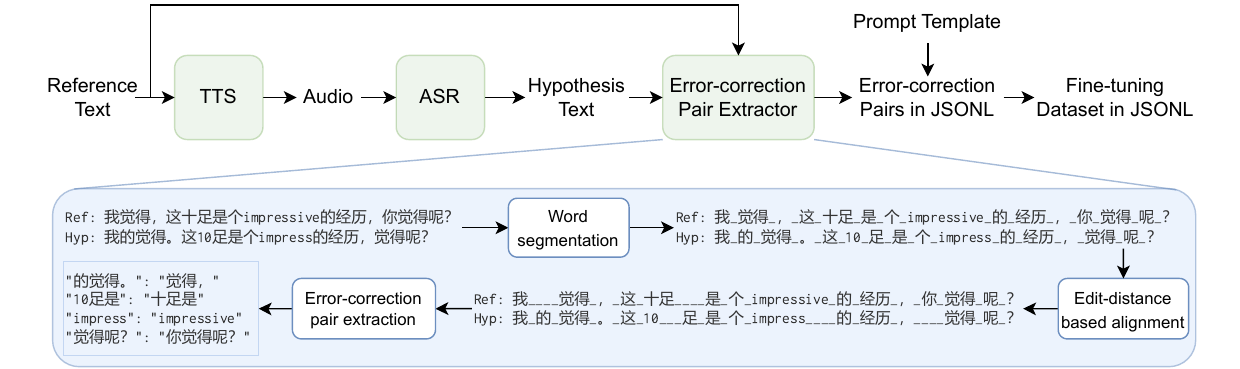}
    \caption{The pipeline of the ChFT dataset creation and fine-tuning dataset preparation.}

    \label{fig:pipeline}
\end{figure*}

\section{Chinese Full-text Error Correction Dataset}
\label{sec:dataset}

The Chinese Full-text Error Correction Dataset (ChFT) is constructed following the pipeline shown in Fig.~\ref{fig:pipeline},
which consists of the following four main steps: text collection, text-to-speech (TTS), ASR, and error-correction pair extractor. The resources used in each step are as follows:
\begin{itemize}
    \item Text collection: We use a portion of the open-sourced Chinese corpus THUCNews~\cite{THUCNews}. THUCNews is based on historical data from the Sina News RSS subscription channels between 2005 and 2011. It includes 740,000 news documents across 14 categories.
    \item TTS: We use ChatTTS~\cite{ChatTTS} developed by the 2noise team to convert text into speech. ChatTTS enables natural and expressive speech synthesis and surpasses most open-source TTS models in terms of prosody.
    \item ASR: The cascaded ASR pipeline\footnote{https://www.modelscope.cn/models/iic/speech\_paraformer-large\_asr\_nat-zh-cn-16k-common-vocab8404-pytorch} including Voice Activation Detection~\cite{zhang2018deep}, Paraformer ASR~\cite{gao2022paraformer}, punctuation restoration~\cite{chen2020controllable} and ITN\footnote{https://github.com/wenet-e2e/WeTextProcessing} is used to transcribe the long-duration speech into text.
    \item Error-correction pair extractor: Word segmentation\footnote{https://github.com/fxsjy/jieba} is firstly performed on both reference and hypothesis texts to produce semantic word units. By aligning the transcribed text with reference using edit distance, error-correction pairs are generated. Specifically, insertion and deletion errors are transformed into substitutions by padding words after the current word. To ensure the uniqueness of errors, their length is set to a minimum of 4. If an error is shorter than 4, it is padded with subsequent words, so is the corresponding correction.
\end{itemize}

The ChFT dataset possesses several unique characteristics that set it apart from existing datasets, including:
\begin{itemize}
    \item ChFT is a full-text corpus, differing from existing datasets that are mostly sentence-level. This allows for an exploration of both full-text and segment-level error correction. It also covers a wide range of domains, including sports, entertainment, home, lottery, real estate, education, fashion, current affairs, horoscopes, games, society, technology, stocks, and finance.
    \item ChFT contains not only errors related to Chinese characters but also errors related to punctuation and ITN, making it an end-to-end error correction dataset.
    \item To evaluate the model's generalization across various dimensions, ChFT comprises 3 types of test sets: a homogeneous test set with same source as the training set, an up-to-date test set containing the latest articles, and a hard test set, which is a variant of the homogeneous test set with additional background babble noise.
\end{itemize}

\begin{table}[htbp]
    \caption{Numbers of articles and segments in the ChFT dataset.}
    \label{tab:dataset-num}
    \resizebox{0.49\textwidth}{!}{
    \begin{tabular}{|l|c|rrr|}
    \hline
    \multirow{2}{*}{} & \multirow{2}{*}{\textbf{Train}}      & \multicolumn{3}{c|}{\textbf{Test}}                                                                      \\ \cline{3-5} 
                      &                             & \multicolumn{1}{l|}{homogeneous} & \multicolumn{1}{l|}{up-to-date} & \multicolumn{1}{l|}{hard} \\ \hline
    No. of articles   & \multicolumn{1}{r|}{41,651}  & \multicolumn{1}{r|}{5,592}       & \multicolumn{1}{r|}{3,987}            & 5,590                      \\ \hline
    No. of segments   & \multicolumn{1}{r|}{441,629} & \multicolumn{1}{r|}{58,920}      & \multicolumn{1}{r|}{13,287}           & 48,212                     \\ \hline
    \end{tabular}
    }
    \end{table}

\begin{figure}[htbp]
        \centering
        \includegraphics[trim=40 0 60 25,clip,width=\linewidth]{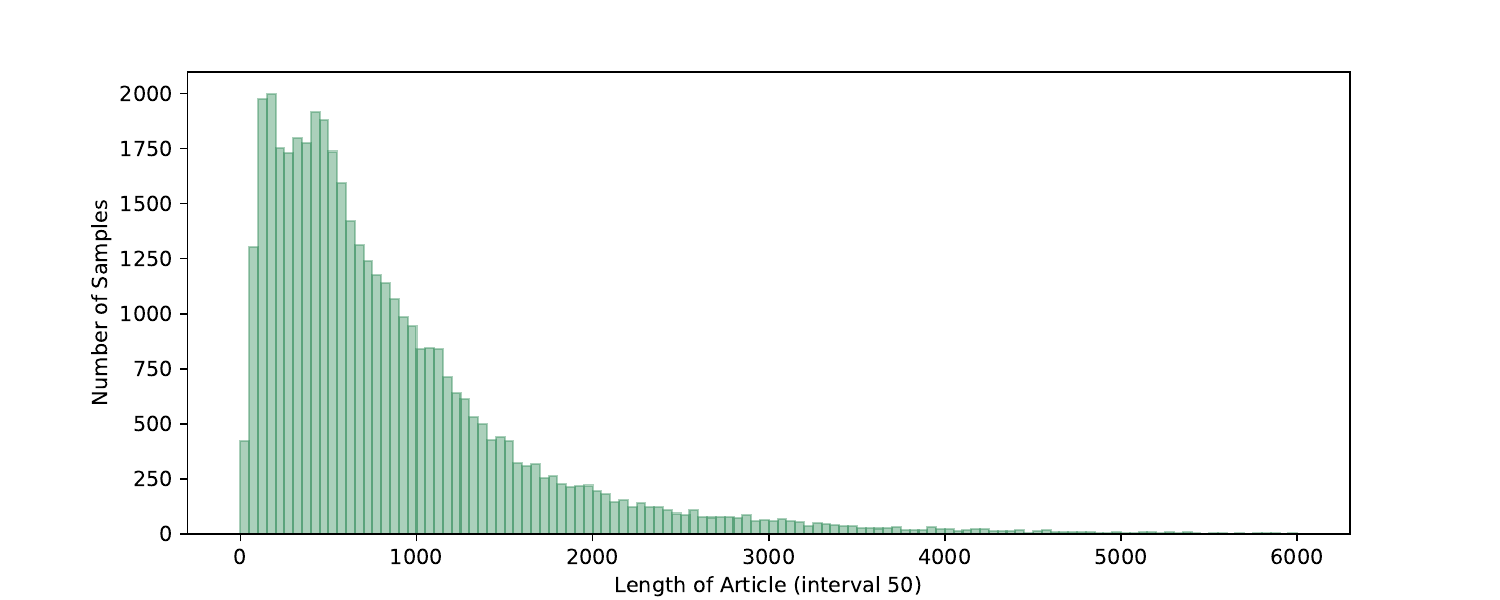}
        \includegraphics[trim=35 0 60 10,clip,width=\linewidth]{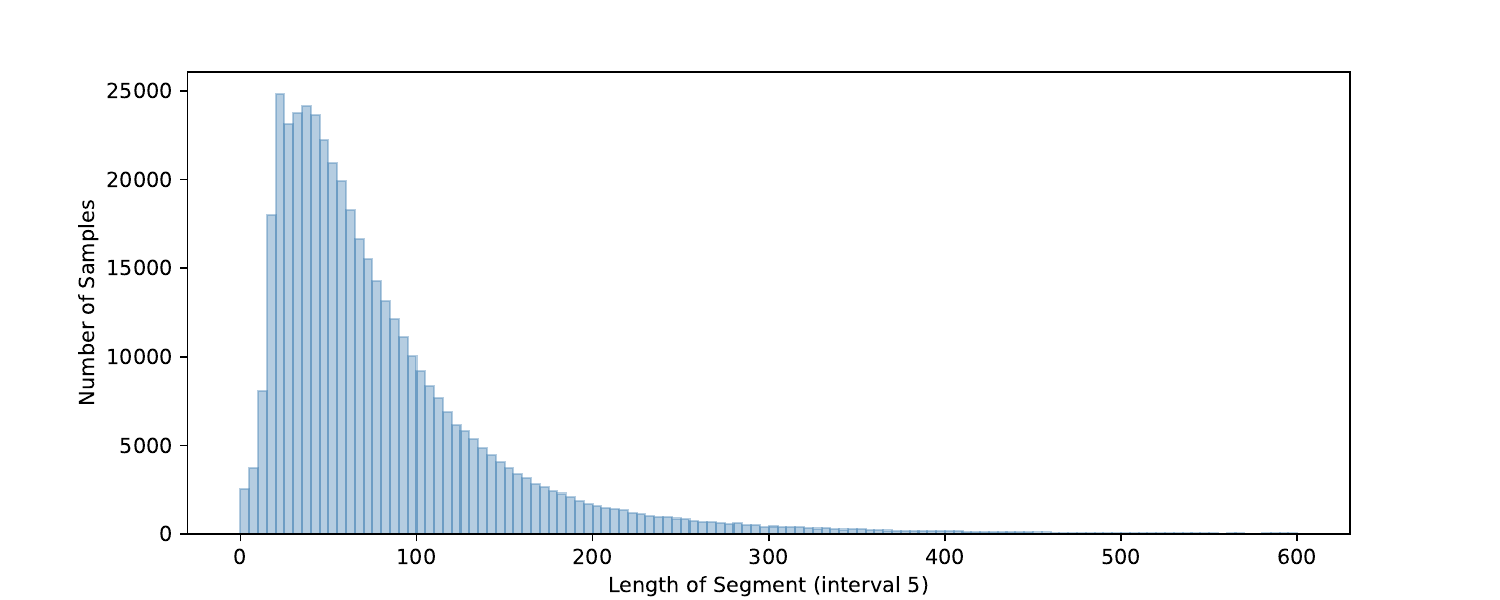}
        \caption{Length distributions for articles (up) and segments (down) in training set.}
        \label{fig:dataset-len}
    \end{figure}

To study the impact of text length on error correction performance with long context, we also split the full-texts into segments, each is composed of several sentences. The segment ends at a point where both reference and hypothesis end with terminal punctuation marks, i.e., ``\begin{CJK*}{UTF8}{gbsn}。\end{CJK*}", ``\begin{CJK*}{UTF8}{gbsn}？\end{CJK*}" and ``\begin{CJK*}{UTF8}{gbsn}！\end{CJK*}".
Table~\ref{tab:dataset-num} shows the numbers of articles and segments in the ChFT dataset for training and testing. 
Fig.~\ref{fig:dataset-len} shows the distribution of the lengths of articles and segments in the training set, where most articles have lengths between 100 and 1000, and most segments have lengths between 20 and 100.
The segments have relatively longer lengths than commonly used sentences in previous studies, providing more context for the model to correct errors.
Shorter sentences with pure text are not studied in this paper.

\section{Prompt Design}
\label{sec:prompt}

We design the prompt based on the following considerations: 1) the length of the text to be corrected, i.e., full-text or segment; 2) the output format of the result from LLM.
For the latter, the first type of output format is the final corrected text itself. Considering that when input text becomes longer, LLM is more likely to hallucinate, and also the direct corrected text lacks controllability on specific errors, we design a JSON format for output to just provide error-correction pairs for each segment in the input text separately. The JSON output format can be applied to both full-text and segment, so we get four types of prompts in total as shown in Table~\ref{tab:prompt}.

\begin{table}[htbp]
    \caption{Different types of prompts.}
    \label{tab:prompt}
    \resizebox{0.49\textwidth}{!}{
    \begin{tabular}{|l|l|l|}
    \hline
    \textbf{Prompt type} & \textbf{Input text type} & \textbf{Output type}             \\ \hline
    article\_direct      & full-text article   & final corrected text             \\ \hline
    article\_json        & full-text article   & error-correction pairs with JSON \\ \hline
    seg\_direct          & multi-sentence segment             & final corrected text             \\ \hline
    seg\_json            & multi-sentence segment             & error-correction pairs with JSON \\ \hline
    \end{tabular}
    }
    \end{table}

    \begin{figure}[htbp]
        \centering
          \includegraphics[trim=545 110 580 5,clip,width=\linewidth]{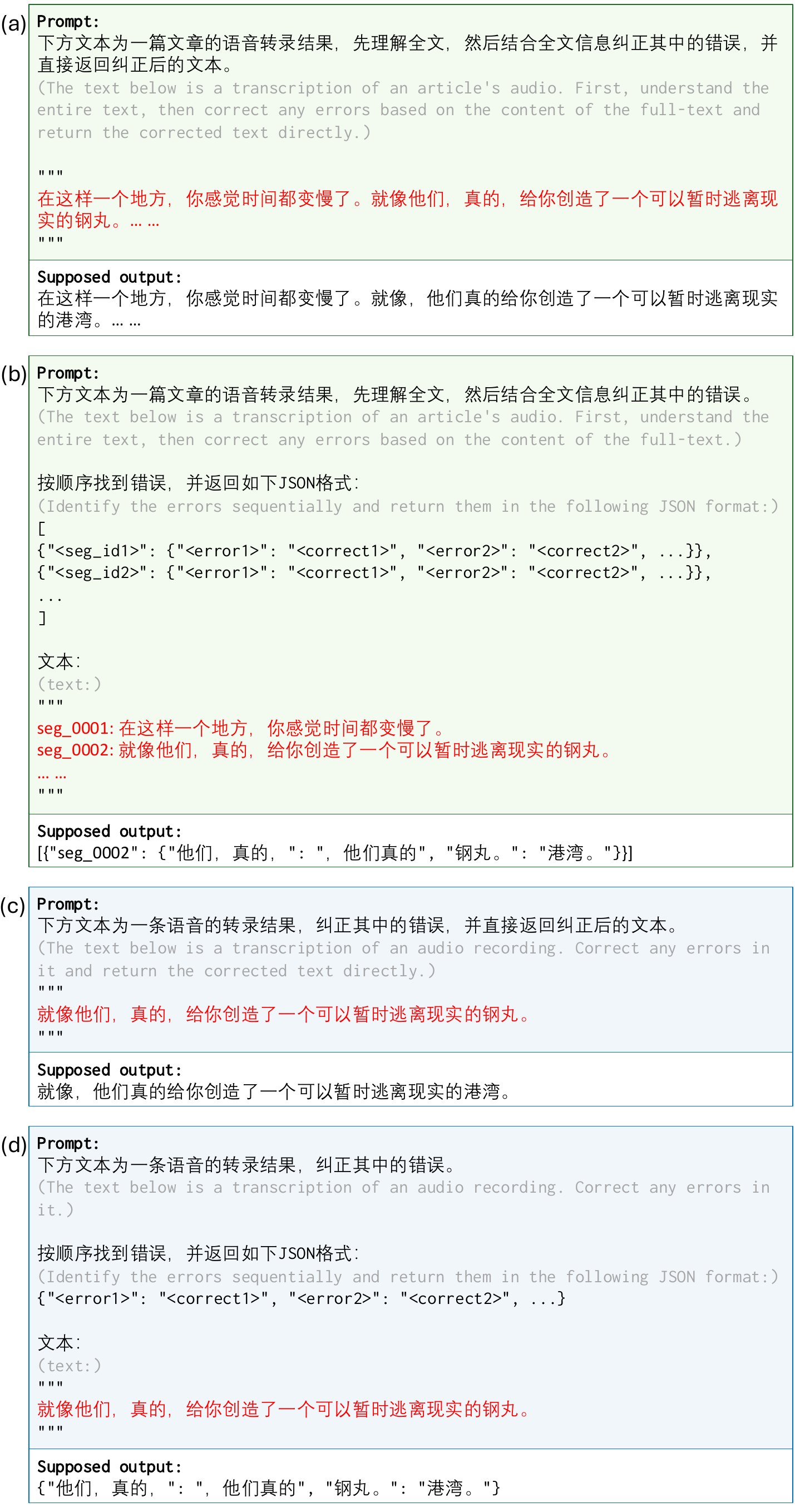}
          \caption{ Examples of different prompts and supposed outputs for error correction: (a) article\_direct, (b) article\_json, (c) seg\_direct and (d) seg\_json.  The {\color{gray}{gray}} part in parentheses is just for translation.
          The {\color{red}{red}} part is the full-text or segment to be corrected.}
      
          \label{fig:prompt}
        \end{figure}

The example for each prompt and supposed output are also shown in Fig.~\ref{fig:prompt}, where a full-text or segment is inserted into the corresponding prompt template to generate the final prompt for the LLM during training and inference.
Note that for the `article\_json' prompt, the full-text is structured as multiple segments for JSON output to better locate errors and each segment has its own dictionary for error-correction pairs, whereas the output of `segment\_json' includes only a single dictionary for the entire segment.


\section{Experiments}
\label{sec:experiments}

We conducted a series of experiments to evaluate the effectiveness of the four prompts presented in Table~\ref{tab:prompt} for fine-tuning ChatGLM~\cite{glm2024chatglm}, specifically the GLM-4-9B-Chat model\footnote{https://huggingface.co/THUDM/glm-4-9b-chat} with 9 billion parameters.

The entire ChFT training set was used for fine-tuning, reserving 10\% of the data for validation. 
Following the steps outlined in Fig.~\ref{fig:pipeline}, we created fine-tuning datasets for each prompt listed in Table~\ref{tab:prompt}.
After separately fine-tuning the model with each prompt, we evaluated the performance of all fine-tuned models on three ChFT test sets.

We utilized 8 NVIDIA A100 GPUs to fine-tune ChatGLM with one epoch for each prompt. To improve computational efficiency, we used the low-rank adaptation (LoRA) method~\cite{hu2021lora} with a rank of 16 on all linear modules, optimizing a limited number of parameters. The fine-tuning pipeline followed the open-source tool LlamaFactory~\cite{zheng2024llamafactory}. 

Following ASR evaluation metric, the Character/Word Error Rate (ER) was employed to assess performance for pure Chinese or code-switched English. The ER is defined as:

{\footnotesize
$$
ER = \frac{S+D+I}{N}
$$
}

\noindent where \( N \) represents the total number of characters/words in reference, and \( S \), \( D \), and \( I \) correspond numbers of substitutions, deletions, and insertions, respectively. For clearer comparison, we also used Error Rate Reduction (ERR):

{\footnotesize
$$
ERR = \frac{ER_{\text{fine-tuned}} - ER_{\text{baseline}}}{ER_{\text{baseline}}}
$$
}

\noindent where \( ER_{\text{fine-tuned}} \) and \( ER_{\text{baseline}} \) are the ERs from the fine-tuned LLM and the baseline ASR, respectively.
For punctuation and ITN correction, we employed the same evaluation method as used in ASR.

\subsection{Homogeneous Test}
The homogeneous test set was sourced from the same dataset, THUCNews, as the training set. Results are shown in Table \ref{tab:res-homogeneous}. The table indicates that all four prompts significantly improve overall performance compared to the baseline ASR.

\begin{table}[htbp]
    \caption{Results of 4 prompts for error correction with fine-tuned ChatGLM on the homogeneous test set.}
    \label{tab:res-homogeneous}
    \resizebox{0.5\textwidth}{!}{
    \begin{tabular}{|l|lllll|}
    \hline
    \multirow{2}{*}{\textbf{Prompt type}} & \multicolumn{5}{c|}{\textbf{ER\%}$\downarrow$ $_{ERR\%\downarrow}$} \\ \cline{2-6} 
                                 & \multicolumn{1}{l|}{Mandarin} & \multicolumn{1}{l|}{Punctuation} & \multicolumn{1}{l|}{ITN} & \multicolumn{1}{l|}{English} & Overall \\ \hline
    Baseline                     & \multicolumn{1}{l|}{6.16}     & \multicolumn{1}{l|}{67.18}       & \multicolumn{1}{l|}{45.17}  & \multicolumn{1}{l|}{80.53}   & 12.61   \\ \hline
    article\_direct              & \multicolumn{1}{l|}{6.51$_{\color{gray} 5.68}$}     & \multicolumn{1}{l|}{37.34$_{\color{teal} -44.42}$}       & \multicolumn{1}{l|}{31.37$_{\color{teal} -30.55}$}  & \multicolumn{1}{l|}{58.65$_{\color{teal} -27.17}$}   & 9.97$_{\color{teal} -20.94}$ \\ \hline
    article\_json                & \multicolumn{1}{l|}{4.78$_{\color{teal} -22.40}$}     & \multicolumn{1}{l|}{37.68$_{\color{teal} -43.91}$}       & \multicolumn{1}{l|}{29.23$_{\color{teal} -35.29}$}  & \multicolumn{1}{l|}{56.41$_{\color{teal} -29.95}$}   & 8.41$_{\color{teal} -33.31}$ \\ \hline
    seg\_direct                  & \multicolumn{1}{l|}{5.33$_{\color{teal} -13.47}$}     & \multicolumn{1}{l|}{32.36$_{\color{teal} -51.83}$}       & \multicolumn{1}{l|}{26.55$_{\color{teal} -41.22}$}  & \multicolumn{1}{l|}{55.27$_{\color{teal} -31.37}$}   & 8.38$_{\color{teal} -33.54}$ \\ \hline
    seg\_json                    & \multicolumn{1}{l|}{5.43$_{\color{teal} -11.85}$}     & \multicolumn{1}{l|}{42.03$_{\color{teal} -37.44}$}       & \multicolumn{1}{l|}{28.16$_{\color{teal} -37.66}$}  & \multicolumn{1}{l|}{56.78$_{\color{teal} -29.49}$}   & 9.33$_{\color{teal} -26.01}$ \\ \hline
    \end{tabular}
    }
    \end{table}

Specifically, for Mandarin errors—which are crucial to the task—the `article\_direct' prompt underperformed compared to the baseline ASR. This may be due to the LLM's tendency to generate hallucinations for lengthy output. Conversely, when the LLM outputs error-correction pairs in JSON format as with the `article\_json' prompt, hallucinations are substantially minimized, leading to the best performance. For segment input texts, both direct output and JSON output demonstrate improvements over the baseline ASR. 
Notably, `seg\_direct,' which provides directly corrected output, outperforms the JSON output, contrary to what is observed with article input.
For other types of errors, such as punctuation, ITN, and code-switched English, all prompts show noticeable corrections.

\subsection{Up-to-date Test}
The reference text for the up-to-date test set was derived from a subset of rthk\_news\footnote{https://huggingface.co/datasets/jed351/rthk\_news}, selecting articles from July 1, 2024, to August 11, 2024, each containing more than 100 characters. Given that the GLM-4-9B-Chat checkpoint was released on June 4, 2024, the LLM had never encountered the reference text for the up-to-date test before. This test set aims to assess the LLM's ability to generalize to entirely novel data. Results are shown in Table \ref{tab:res-up-to-date}. The table demonstrates that the correction performance of the LLM on the most recent test set aligns with the trends observed in the homogeneous test set, indicating robust generalization capabilities on the latest data.

\begin{table}[htbp]
    \caption{Results of 4 prompts for error correction with fine-tuned ChatGLM on the up-to-date test set.}
    \label{tab:res-up-to-date}
    \resizebox{0.5\textwidth}{!}{
    \begin{tabular}{|l|lllll|}
    \hline
    \multirow{2}{*}{\textbf{Prompt type}} & \multicolumn{5}{c|}{\textbf{ER\%}$\downarrow$ $_{ERR\%\downarrow}$} \\ \cline{2-6} 
                                 & \multicolumn{1}{l|}{Mandarin} & \multicolumn{1}{l|}{Punctuation} & \multicolumn{1}{l|}{ITN} & \multicolumn{1}{l|}{English} & Overall \\ \hline
    Baseline                     & \multicolumn{1}{l|}{5.97}     & \multicolumn{1}{l|}{55.68}       & \multicolumn{1}{l|}{38.23}  & \multicolumn{1}{l|}{89.85}   & 10.80    \\ \hline
    article\_direct              & \multicolumn{1}{l|}{6.21$_{\color{gray} 4.02}$}     & \multicolumn{1}{l|}{29.59$_{\color{teal} -46.86}$}       & \multicolumn{1}{l|}{20.88$_{\color{teal} -45.38}$}  & \multicolumn{1}{l|}{53.01$_{\color{teal} -41.00}$}   & 8.48$_{\color{teal} -21.48}$ \\ \hline
    article\_json                & \multicolumn{1}{l|}{5.13$_{\color{teal} -14.07}$}     & \multicolumn{1}{l|}{32.78$_{\color{teal} -41.13}$}       & \multicolumn{1}{l|}{19.14$_{\color{teal} -49.93}$}  & \multicolumn{1}{l|}{52.38$_{\color{teal} -41.70}$}   & 7.75$_{\color{teal} -28.24}$ \\ \hline
    seg\_direct                  & \multicolumn{1}{l|}{5.04$_{\color{teal} -15.58}$}     & \multicolumn{1}{l|}{24.78$_{\color{teal} -55.50}$}       & \multicolumn{1}{l|}{21.64$_{\color{teal} -43.40}$}  & \multicolumn{1}{l|}{53.47$_{\color{teal} -40.49}$}   & 7.04$_{\color{teal} -34.81}$ \\ \hline
    seg\_json                    & \multicolumn{1}{l|}{5.29$_{\color{teal} -11.39}$}     & \multicolumn{1}{l|}{35.19$_{\color{teal} -36.80}$}       & \multicolumn{1}{l|}{18.20$_{\color{teal} -52.39}$}   & \multicolumn{1}{l|}{52.69$_{\color{teal} -41.36}$}   & 8.08$_{\color{teal} -25.19}$ \\ \hline
    \end{tabular}
    }
    \end{table}

\subsection{Hard Test}
Given the relatively high baseline performance of the above test sets, we also conducted tests on a hard test set, derived from the homogeneous test set with the addition of background babble noise. Results are shown in Table \ref{tab:res-hard}. The overall performance of all prompts declined due to poorer correction of Mandarin. This deterioration likely stems from the background noise complicating transcription and introducing more Mandarin errors. Nonetheless, with error-correction JSON output, the LLM model still achieved consistent improvements. Other types of errors were still effectively corrected.

\begin{table}[htbp]
    \caption{Results of 4 prompts for error correction with fine-tuned ChatGLM on the hard test set.}
    \label{tab:res-hard}
    \resizebox{0.5\textwidth}{!}{
    \begin{tabular}{|l|lllll|}
    \hline
    \multirow{2}{*}{\textbf{Prompt type}} & \multicolumn{5}{c|}{\textbf{ER\%}$\downarrow$ $_{ERR\%\downarrow}$} \\ \cline{2-6} 
                                 & \multicolumn{1}{l|}{Mandarin} & \multicolumn{1}{l|}{Punctuation} & \multicolumn{1}{l|}{ITN} & \multicolumn{1}{l|}{English} & Overall \\ \hline
    Baseline                     & \multicolumn{1}{l|}{19.77}    & \multicolumn{1}{l|}{70.55}       & \multicolumn{1}{l|}{54.89}  & \multicolumn{1}{l|}{89.08}   & 25.24   \\ \hline
    article\_direct              & \multicolumn{1}{l|}{24.79$_{\color{gray} 25.39}$}    & \multicolumn{1}{l|}{51.44$_{\color{teal} -27.09}$}       & \multicolumn{1}{l|}{49.67$_{\color{teal} -9.51}$}  & \multicolumn{1}{l|}{75.16$_{\color{teal} -15.63}$}   & 27.89$_{\color{gray} 10.50}$ \\ \hline
    article\_json                & \multicolumn{1}{l|}{18.85$_{\color{teal} -4.65}$}    & \multicolumn{1}{l|}{49.80$_{\color{teal} -29.41}$}        & \multicolumn{1}{l|}{45.00$_{\color{teal} -18.02}$}  & \multicolumn{1}{l|}{71.99$_{\color{teal} -19.19}$}   & 22.36$_{\color{teal} -11.41}$ \\ \hline
    seg\_direct                  & \multicolumn{1}{l|}{20.21$_{\color{gray} 2.23}$}    & \multicolumn{1}{l|}{43.57$_{\color{teal} -38.24}$}       & \multicolumn{1}{l|}{42.46$_{\color{teal} -22.65}$}  & \multicolumn{1}{l|}{71.58$_{\color{teal} -19.65}$}   & 22.99$_{\color{teal} -8.91}$ \\ \hline
    seg\_json                    & \multicolumn{1}{l|}{19.48$_{\color{teal} -1.47}$}    & \multicolumn{1}{l|}{52.32$_{\color{teal} -25.84}$}       & \multicolumn{1}{l|}{42.60$_{\color{teal} -22.39}$}  & \multicolumn{1}{l|}{72.67$_{\color{teal} -18.42}$}   & 23.09$_{\color{teal} -8.52}$ \\ \hline
    \end{tabular}
    }
    \end{table}

\subsection{Limitations}
\label{sec:limitation}
In this study, all audio data was machine-generated, which does not reflect enough real-world conditions. Consequently, the error pattern of ASR and correction performance of the fine-tuned LLM may not be representative of actual scenarios. Future work will involve experiments using real-world audio data to evaluate the fine-tuned LLM's performance.

\section{Conclusion}
\label{sec:conclusion}
This paper explores the ability of LLM to correct Chinese ASR errors in terms of text, punctuation and ITN within full-text. To this end, the ChFT dataset was created and is publicly accessible. Four different prompts were designed for error correction using LLM, considering the input text scales as either article or segment and the output format as either direct correction or error-correction JSON. Fine-tuning ChatGLM with the ChFT dataset under various prompts revealed that the LLM effectively corrects ASR errors in full-text contexts. In future research, we plan to explore long audios in real-world scenarios and devise more advanced prompts that incorporate additional context information, such as hot words, to enhance the LLM's correction performance.

\bibliographystyle{IEEEtran}
\bibliography{mybib}

\end{document}